%% file: acl_latex.tex
\documentclass[11pt]{article}

\usepackage[preprint]{acl}

\usepackage{times}
\usepackage{latexsym}

\usepackage[T1]{fontenc}

\usepackage[utf8]{inputenc}

\usepackage{microtype}

\usepackage{inconsolata}

\usepackage{graphicx}
\usepackage{xspace}
\usepackage{tcolorbox}
\usepackage{soul}
\usepackage{booktabs}
\usepackage[dvipsnames]{xcolor}
\usepackage{multirow}
\newtcolorbox{takeaway}{
  colback=sysBlue!5,
  colframe=sysBlue,
  fonttitle=\bfseries\small,
  boxsep=2pt, left=4pt, right=4pt, top=2pt, bottom=2pt, toptitle=1pt, bottomtitle=1pt, before skip=8pt, after skip=0pt
}
\newcommand{\un}{\underline}

\input{gradient_font}
\newcommand{\sysn}{\textsc{\gradienttext{sysBlue}{sysGreen}{HOTFIXR}}\xspace}

\title{LLMs Get Smarter from Targeted Synthetic Multilingual Data}

\author{
\textbf{Ishika Agarwal\textsuperscript{1}},
\textbf{Arkajyoti Chakraborty\textsuperscript{2}},
\textbf{Tanner Sorensen\textsuperscript{2}},
\\
\textbf{Neha Gupta\textsuperscript{2}},
\textbf{Andreas Stolcke\textsuperscript{2}}
\\
\\
 \textsuperscript{1}UIUC,
 \textsuperscript{2}Uniphore
\\
 \small{
   \textbf{Correspondence:} \href{mailto:ishikaa2@illinois.edu}{ishikaa2@illinois.edu}
 }
}

\definecolor{cBase}{RGB}{154,157,163}
\definecolor{cEngReason}{RGB}{255,127,14}
\definecolor{cSelGT}{RGB}{140,86,75}
\definecolor{cSelGEN}{RGB}{148,103,189}
\definecolor{cFiltered}{RGB}{213,125,191}
\definecolor{cUntrained}{RGB}{242,193,78}
\definecolor{cDataEnvGym}{RGB}{214,39,40}

\newcommand{\Untrained}{\sethlcolor{cUntrained}\hl{Untrained}\xspace}

\newcommand{\Base}{\textcolor{black}{Base}\xspace}
\newcommand{\EngReason}{\textcolor{cEngReason}{EngReason}\xspace}
\newcommand{\SelGT}{\textcolor{cSelGT}{SelectionGT}\xspace}
\newcommand{\SelGEN}{\textcolor{cSelGEN}{SelectionGEN}\xspace}
\newcommand{\Filtered}{\textcolor{cFiltered}{Filtered}\xspace}
\newcommand{\DataEnvGym}{\textcolor{cDataEnvGym}{DataEnvGym}\xspace}
\newcommand{\plus}{\textcolor{ForestGreen}{+}}
\newcommand{\minus}{\textcolor{BrickRed}{-}}

\begin{document}
\maketitle
\begin{abstract}
Language-specific competency (LSC) is the phenomenon of a language model performing better or worse depending on the language of the prompt. In other words, a language model outputs different (and potentially incorrect) responses to the same semantic query when prompted in different languages. Prior work attributes this to an internal misalignment of semantic representation across languages. Currently, there are two main approaches to address LSC in the literature: (1) routing all queries through English, improving performance, but limiting language expressivity to English; or (2) training on language-balanced data, equalizing model performance across languages, but reducing overall performance. In this work, we take a data centric perspective and introduce \sysn: \textbf{H}ardness \textbf{O}ptimized \textbf{T}raining-data \textbf{F}or \textbf{I}mproving \textbf{X}-lingual \textbf{R}easoning. It is a data generation framework that uses models to probe and learn a student model's multilingual weaknesses, and generates data to mitigate them. \sysn can generate multilingual synthetic training data that can improve multilingual performance. We evaluate on three in-distribution tasks, three out-of-distribution tasks, and four out-of-distribution languages. On average, \sysn (1) improves in-distribution performance by 6.2\%, (2) reduces catastrophic forgetting (induced by fine-tuning) on OOD tasks by 3.7\%, and (3) on OOD languages by 7.1\%. Overall, as many real-world applications requires multilingual LLMs, our work contributes to the efforts of making LLMs multilingually proficient. We will release code upon acceptance.
\end{abstract}

\begin{table}[h]
\centering
\small
\begin{tabular}{lccc}
\toprule
 & \multicolumn{2}{c}{Performance} & Language \\ \cmidrule{2-3}
Method & ID ($\uparrow$) & OOD ($\uparrow$) & Spread ($\downarrow$)\\
\midrule
\Base & \un{51.9} & \textbf{65.6} & 9.6 \\
\EngReason & 50.8 & \un{65.4} & 8.8 \\
\SelGT & 49.2 & 60.9 & 9.8 \\
\SelGEN & 50.2 & 55.8 & \un{9.5} \\
\Filtered & 48.9 & 58.2 & 10.6 \\
\Untrained & 50.4 & 63.4 & 9.7 \\
\DataEnvGym & 49.0 & 57.4 & 10.2\\
\midrule
\sysn (ours) & \textbf{56.2} & 64.7 & \textbf{9.4} \\
\bottomrule
\end{tabular}
\caption{\textbf{\sysn has the best in-distribution performance, the best out-of-distribution performance among the training-based baselines, and remains multilingually consistent.} This table shows the cross-lingual consistency with respect to ID and OOD performance. Language spread is the standard deviation of performance across languages. Low spread indicates more consistent performance across languages. Note: the training-based baselines are \SelGT, \SelGEN, \Filtered, \Untrained, and \DataEnvGym.}
\label{tab: spread_summary}
\end{table}

\section{Introduction}
\begin{figure}
    \centering
    \includegraphics[width=0.8\linewidth]{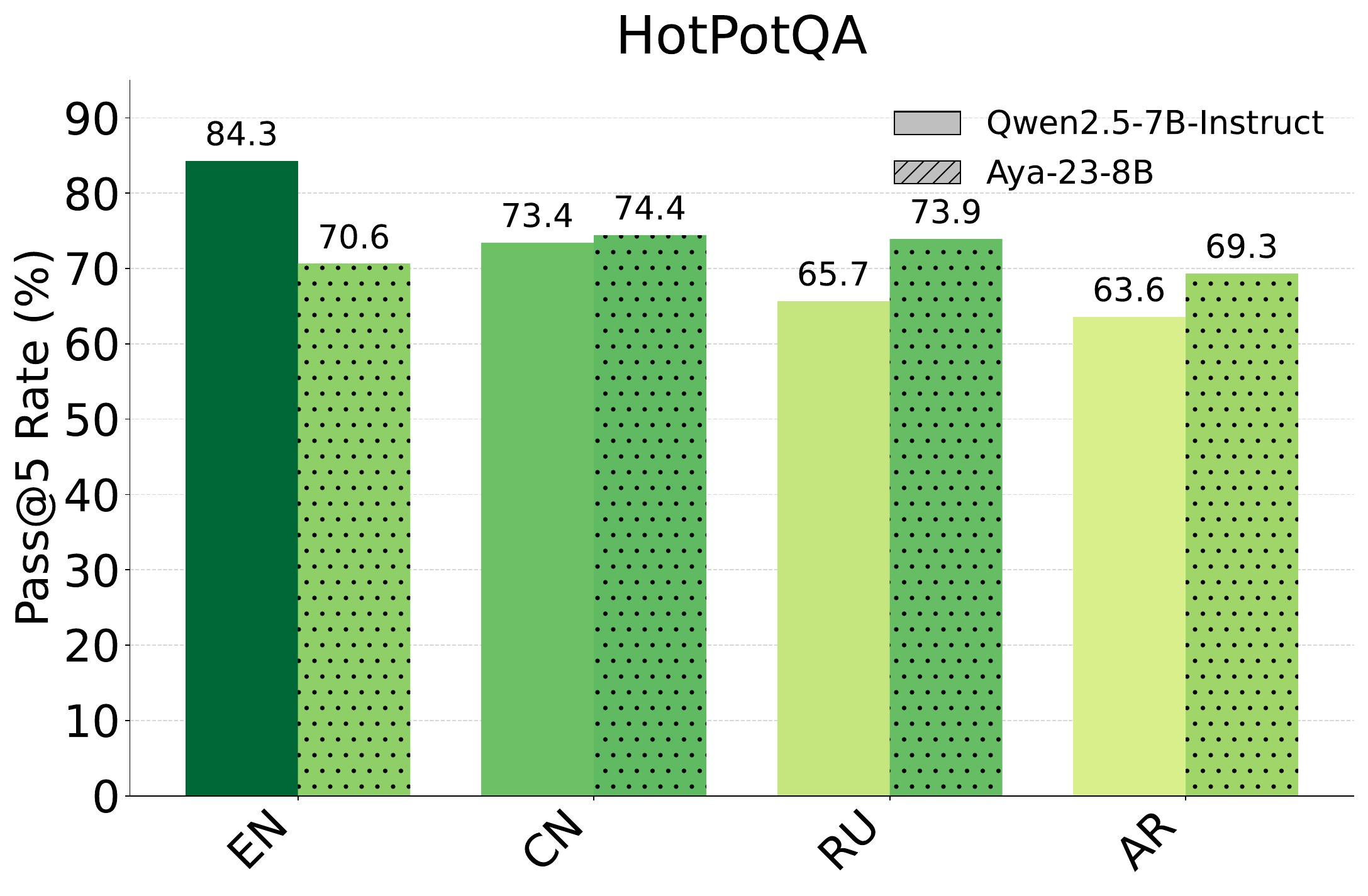}
    \caption{Motivating results for LSC: the performance of Aya (with a more language-balanced pretraining dataset) versus Qwen (with a more English-dominant pretraining dataset) on multilingual HotPotQA \citep{hotpotqa}. We see that although Aya performs more consistently across languages, Aya degrades in performance (see Qwen's performance in English).}
    \label{fig: aya_vs_qwen}
\end{figure}

Many LLM-related business use cases require models to reliably converse in languages other than English. Multilingual language models are trained by adding in non-English data \citep{aya_dataset}. However, English data significantly dominates the training data. As a result, a lot of internal LLM processes happen in English \citep{wendler-etal-2024-llamas}. Specifically, as an LLM processes Spanish (or any non-English) text, it tries to anchor its understanding in English, before answering in Spanish. This anchoring could upper-bound what an LLM can understand in non-English. This points to the representational gap in LLMs, where they represent the same concept differently, in different languages. We also see empirical evidence of this. In Figure \ref{fig: aya_vs_qwen}, we evaluate \texttt{Qwen/Qwen2.5-7B-Instruct} \citep{qwen2.5} on multilingual HotPotQA \citep{hotpotqa} (a multilingual RAG dataset). This plot shows that the model capability varies by language.

If the discrepancy was created by imbalanced data, what happens when we balance the pretraining data? \citet{aryabumi2024aya} trained \texttt{Cohere/aya-23-8B} with a focus on balancing 23 languages in their pretraining data as much as possible. In Figure \ref{fig: aya_vs_qwen}, we also included the performance of this Aya model.

With the language-balanced pretraining dataset, Aya is able to perform more consistently across languages. However, it doesn't perform as well as Qwen does on English. Of course, the models differ in model architecture, ingested data, training strategies, and more, making this more illustrative rather than controlled. Still, this pattern matches the lessons from prior work: \citet{conneau-etal-2020-unsupervised}'s ``Curse of Multilinguality'' for unsupervised learning states that as more languages are added, the overall performance plateaus. \textit{Overall, we are seeing a pattern of degradation when models are trained with multilingual data.}

To summarize, the multilingual tradeoff is to either optimize for English performance and suffer in non-English tasks, or to optimize for consistency in English/non-English performance and suffer in overall model accuracy. We try to explore a middle ground: if data is truly that important, can we curate post-training data to teach a model perform well both across languages and overall?

We introduce \sysn: a post-training data generation framework that aims to improve an LLM's multilingual ability without degrading its general abilities. \sysn is a synthetic data generation framework that trains a question generation model based on a student model's feedback signals. The question generator is trained to probe the student's weakness (Section \ref{sec: method}). We evaluate \sysn across a variety of baselines and benchmarks to support our claims (Section \ref{sec: experiments}).

\section{Related Works}
Multilingual LLMs try to map the same semantic phrases into one unified representation space \citep{Xu_2025, ghosh2025surveymultilingualreasoninglanguage}. However, representation misalignments occur within languages themselves. This usually emerges in three behaviors: (1) language-specific knowledge \citep{agarwal2026languagespecificknowledgemodels} where a model can output a different answer to the same factual query in different languages, (2) non-isomorphism \citep{wu-etal-2024-representational} where certain words or phrases don’t have direct translations in other languages, (3) non-compositionality \citep{agarwal2026risingtideliftsboats} where the meaning cannot be derived (and therefore, translated) from individual words, including idioms and metaphors.

Previous work has shown this misalignment is mainly due to the English-dominant pretraining data \citep{aryabumi2024aya}. This gears the model towards internally representing language in English \citep{wendler-etal-2024-llamas, schut2025multilingualllmsthinkenglish, zhao_multilingual}. Particularly, \citet{wendler-etal-2024-llamas} show that models route all their latent thinking into English spaces in the early-to-middle layers, then route their thinking back to non-English spaces in the later layers. This results in English becoming an upper bound for representation: concepts that can be represented in English are more likely to result in accurate answers. The key is to realign these concepts; Section 3.2 shows evidence that they can be realigned in pretraining, but pretraining is prohibitively expensive in many settings.

\paragraph{Mitigating misalignment} Misalignment can be mitigated using a few strategies. First there is smart prompting. Researchers craft particular prompts that elicit language-dependent knowledge from a model to increase performance on a multi-lingual task \citep{donthi-etal-2025-improving}. The limitation of this approach is that the prompts are hand-crafted and might not generalize across language models. Models can also be sensitive to the style of the prompt, making this an unreliable method. Next is contrastive learning. Usually, a contrastive learning objective is employed to re-align representations \citep{tan-etal-2023-multilingual, li-etal-2024-improving-cross-lingual, zhang-etal-2026-speak} and requires positive-negative paired data in order to improve the representations. This is more reliable than prompting, but still requires high-quality samples with human ground truth.

Focusing on multilingual reasoning specifically, one way to mitigate is to make architectural changes. For example, \citet{yoon-etal-2024-langbridge} uses a multilingual model to provide embedding inputs to a reasoning model to use the specialized capabilities of each model. Other fine-tuning methods either try to align low-resource language reasoning chains with those from high-resource languages \citep{she-etal-2024-mapo} or carefully try to craft correct reasoning traces in other languages to improve the performance \citep{ranaldi-pucci-2025-multilingual}. Finally, other works also focus on data curation for aligning representations.

\paragraph{Multilingual data curation.} Most of the data curation works are focused on filtering and cleaning up existing datasets \citep{nguyen-etal-2024-culturax, li-etal-2024-x, penedo2025fineweb, messmer-etal-2025-enhancing}. The cleaning strategies involve deduplication and removing multilingual documents. The filtering strategies involve model-based data selection (keeping training samples that are within the distribution of the existing reference set). \citet{li-etal-2024-x} use a form of self-distillation where they generate cross-lingual samples (instruction in one language, but response in a low-resource language). This helps to generate QA pairs with fewer translation mistakes.

\section{\sysn Methodology}
\label{sec: method}

\begin{figure}[h]
    \centering
    \includegraphics[width=0.8\linewidth]{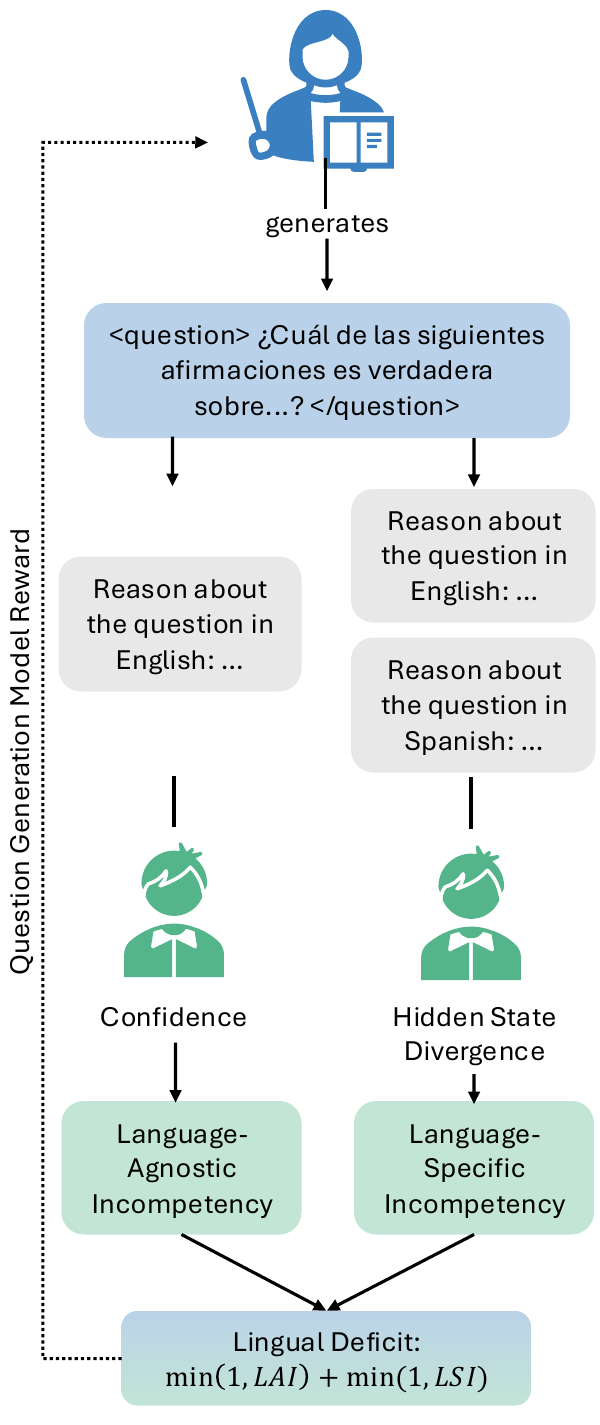}
    \caption{An illustration of how to train the question generation model in \sysn, along with the reward design.}
    \label{fig: method}
\end{figure}

\sysn is a synthetic data generation framework in which we train a question generation model based on the student model's signals. Our framework is inspired by AcquisitionSynthesis \citep{acquisyn}. First, a question generation model is asked to generate a data sample.\footnote{The question generation model is given a prompt that outlines the data sample requirements and we even provide an in-context sample. Prompts are available in Appendix \ref{app: prompts}.} Second, the student model is provided with the data sample, where we design an acquisition function\footnote{Acquisition functions are primarily known as the selection criteria within data selection and active learning works \citep{settles2012active}. We use the term ``acquisition function'' because it is a metric that informs us how good our data is for training purposes.} that indicates whether the sample uncovers a student model's \textit{\textbf{lingual deficit}}. The acquisition function is described in Section \ref{sec: acquisition_function}, but is essentially a difficulty and misalignment score that indicates whether the sample elicits a student model weakness. Finally, this score is given as feedback to the question generation model, where it will be optimized to generate samples that maximize the difficulty score. The optimization algorithm is GRPO \citep{grpo}. Figure \ref{fig: method} contains a visualization of \sysn.

\begin{figure}[h]
    \centering
    \includegraphics[width=\linewidth]{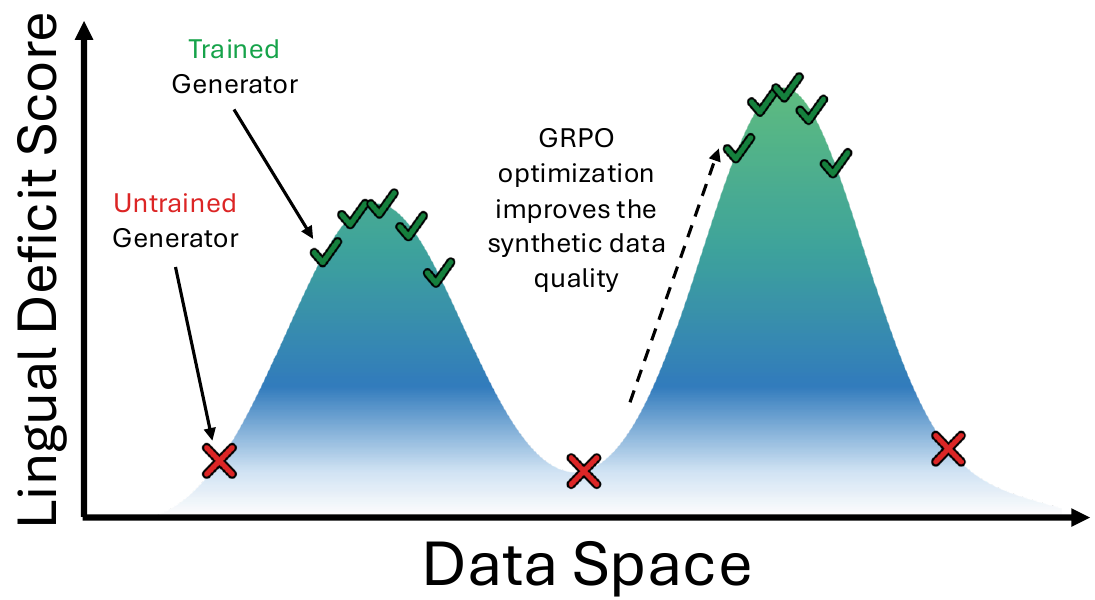}
    \caption{Intuition behind \sysn. The ``data space'' represents the data samples generated by the question generator. Essentially, we want to hill climb through the data space to find samples that have high lingual deficit scores, as they will be most informative to the student model. By fine-tuning the question generation model to generate samples according to the student model's lingual deficit, we can improve the information embedded within data, improving our models on downstream tasks.}
    \label{fig: acquisition}
\end{figure}

\subsection{Measuring Lingual Deficit}
\label{sec: acquisition_function}
Student models have two kinds of weaknesses: language-agnostic incompetencies (cannot answer the question correctly in any language) and language-specific incompetency (cannot answer the some questions correctly in certain languages). We need a metric that can capture both weaknesses.

To measure \textbf{language-agnostic incompetency (LAI)}, we use the LLM's uncertainty in answering the question by using English-centric reasoning. Once the student model receives the question generated by the question generation model, we prompt the student model to ``answer the following question, reason using English''. Given a prompt $P$ and response $R$ (where $R_t^{(i)}$ is the $i$-th most probable token in the $t$-th spot), the uncertainty of a model is $U(P, R) = 1-\Bigl( \frac{1}{|R|} \textstyle\sum_t p_\theta(R_t^{(1)} \mid P, R_{<t})\Bigr)$ or 1 minus the probability of the most probable token, averaged across the sequence. A higher reward is assigned to the question generator if \textit{the student model is uncertain about the question in its strongest language\footnote{We assume this to be English, because the majority of pre-training data is in English.}}.

To measure \textbf{language-specific incompetence (LSI)}, we prompt the student model to answer the same model-generated question in two ways: one with ``answer the following question, reason using English'' ($R_{EN}$), and the other with ``answer the following question, reason using $L$'' ($R_{L}$), where $L$ alternates between French, Spanish, Arabic, Portuguese, or Italian. Next, we find the last reasoning token of both $R_{EN}$ and $R_{L}$ (the token right before the answer delimiter), extract the model's hidden state of the last layer for that token, and compute the cosine distance between them. \textit{The question generator is assigned a higher reward if the distance is larger between the two.} Since the correct answer is invariant to the reasoning language, the two reasoning trajectories should be represented in nearby regions of the representation space. If they differ, this represents a cross-lingual misalignment \citep{wendler-etal-2024-llamas} that the student needs to be trained to reduce.

\paragraph{Intuition.} A question generation model that is trained on a student model's feedback signals learns to generate data that is \textit{good} for the student to learn from. Figure \ref{fig: acquisition} is a schematic of an arbitrary "data" space (what the question generation model can generate) versus the score from the acquisition function. During GRPO optimization, the question generation model figuratively probes the data space to understand what data elicits the highest peak in the acquisition function.

In this case, ``good'' for the student is where the reasoning diverges: either because of incorrect answers or because of the representational gap between languages. Of course, there will always be some difference in the representation of two languages, but the aim is to minimize it so that models can generalize well to other languages without requiring training data in those languages.

\subsection{Training Setup} To train our question generator model, we use 500 samples from the training dataset $\mathcal{D}_{QG}$. These 500 samples are used as in-context samples to inform the data sample generation during rollouts. Our training size is small because we see that with more training samples, models reward hack the student feedback signal, and start to generate the one sample that elicits a local maximum reward. Empirically, 500 samples are a sweet spot between maximizing reward and avoiding reward hacks (more in Appendix \ref{app: reward_hacking}).

After training our question generation model, we use it to create a dataset for a student model, $\mathcal{D}_S$, of 5,000 samples. In both the training dataset $\mathcal{D}_{QG}$ and generated dataset $\mathcal{D}_S$, we prompt the model to generate a sample in a particular language between English, French, Spanish, Arabic, Portuguese, and Italian. To obtain reliable answers for $\mathcal{D}_S$, we use \texttt{Qwen/Qwen2.5-32B-Instruct} \citep{qwen2.5} as our label generation model. We train the student model on $\mathcal{D}_S$ with SFT. Then, we evaluate the student model. The better the student model, the higher quality $\mathcal{D}_S$ is.

\section{Experiments}
\label{sec: experiments}
In this section, we describe our thorough evaluations. 

\subsection{Setup}
For \sysn question generation training set $\mathcal{D}_{QT}$, we use $\sim$500 samples of Nemotron-PTDv2's STEM, MATH, and CHAT subsets (167 samples each). We generate $\mathcal{D}_S$ based on the Nemotron data as well. We use the student models trained on these data to showcase the benefits of our method. 

\paragraph{Models.} We use three models for this task: \texttt{Qwen/Qwen2.5-7B-Instruct} and \texttt{Qwen/Qwen2.5-14B-Instruct} \citep{qwen2.5}, and \texttt{meta-llama/Llama-3.1-8B-Instruct} \citep{llama}. We chose this set to showcase the robustness of our method across model families and model sizes. All question generation models and student models are different instantiations of the same instruction-tuned model. In other words, a slightly different Qwen model will teach itself.

\paragraph{Tasks.} We break up our evaluation into four tasks: (1) \textbf{Nemotron} \citep{NemotronPostTrainingDatasetV2} for agentic reasoning, (2) \textbf{factual} parametric knowledge, tested with MMMLU \footnote{https://huggingface.co/datasets/openai/MMMLU}, (3) \textbf{RAG}-based reading comprehension, tested with mHotPotQA \citep{hotpotqa}, and (4) \textbf{translation}, tested with OPUS-100 \citep{opus}. These four tasks are used to determine whether \sysn can teach models that do not forget previously trained information, as well as improve the multilingual performance of LLMs.

\paragraph{Languages.} As mentioned before, \sysn will generate data in English, French, Spanish, Arabic, Portuguese, and Italian. These are our in-distribution languages -- our training datasets are equally balanced among all six languages. In addition, we use German and Japanese as our out-of-distribution languages, to ensure the performance does not degrade across languages.

\paragraph{Metrics.} Each task requires a different kind of similarity metric. Nemotron STEM and MMMLU are multiple-choice question answering tasks. Hence, these tasks are measured by \textbf{Accuracy}, or the percentage of test samples where the LLM chose the same answer as the ground truth. mHotPotQA (multilingual HotPotQA) has short answers, so we use \textbf{ROUGE}-L to capture the lexical agreement by measuring the n-gram overlap between the few-worded answers. OPUS-100 also has short phrases, but could require a metric to reflect the semantic similarity. Hence, we use an LLM-as-a-Judge \textbf{(LAJ)} to determine on a scale of 1-5 whether an English sentence and the translated version match semantically. We use Prometheus-7b-V2.0 \citep{kim2024prometheus} as our LAJ. Finally, we have two other in-distribution tasks that require semantic similarity for evaluation: Nemotron MATH and Nemotron CHAT. We also use LAJ to evaluate these. A prediction with at least a score of 4/5 is considered correct.

To make the metrics comparable, we discretize them into binary correct/incorrect labels, and report the \% of predictions that are correct. Accuracy is already the binary scale. A prediction with at least an 80\% ROUGE score or at least 4/5 LAJ score is considered correct.


\paragraph{Baselines.} To ensure our method is competitive and comparable, we adopt a variety of baselines to prove the various aspects of our claims.

\begin{enumerate}
    \item \textbf{\Base}: the performance of the base models.
    \item \textbf{\EngReason}: training-free baseline that measures the performance of base models when the model is prompted to ``reason about the question in English''. This baseline represents the strategy of falling back on English as the ``language of thought''.
    \item \textbf{\SelGT}: this is a \textit{data selection} baseline. We use 5,000 samples (with their questions, and \textbf{g}round \textbf{t}ruth answers) from the Nemetron PTDv2STEM, MATH, and CHAT subsets to train a student model. The data selection baselines help evaluate data synthesis compared to just using the available data.
    \item \textbf{\SelGEN}: this is also a \textit{data selection} baseline. The only difference between this and SelectionGT is where the labels come from. SelectionGEN \textbf{gen}erates the labels using \texttt{Qwen/Qwen2.5-32B-Instruct}, instead of obtaining them from the original dataset, and helps isolate the role of ground truth versus generated labels.
    \item \textbf{\Filtered}: this is another \textit{data selection} baseline where we use the Lingual Deficit score to rank 10,000 samples. We select 5,000 samples with the highest Lingual Deficit score.
    \item \textbf{\Untrained}: this is a \textit{data synthesis} baseline. We use the base, untrained model as question generation models to generate datasets for student models. This clarifies the importance of optimizing the data generation models.
    \item  \textbf{\DataEnvGym}: this is another \textit{data synthesis} baseline from \cite{dataenvgym}. They use a teacher model to synthesize data based on a student's weaknesses, irrespective of language. To adapt it to our setting, we add to their prompts to generate data in a particular language (same as ours), and we keep the teacher and student models the same as in our settings.
\end{enumerate}

\subsection{Results}
\textbf{All of the results reported in this section are averaged across 3 runs.} We've only reported the averages in the main plots. To see the error bars, please refer to Appendix \ref{app: per_language}.

\begin{figure}[h]
    \centering
    \includegraphics[width=\linewidth]{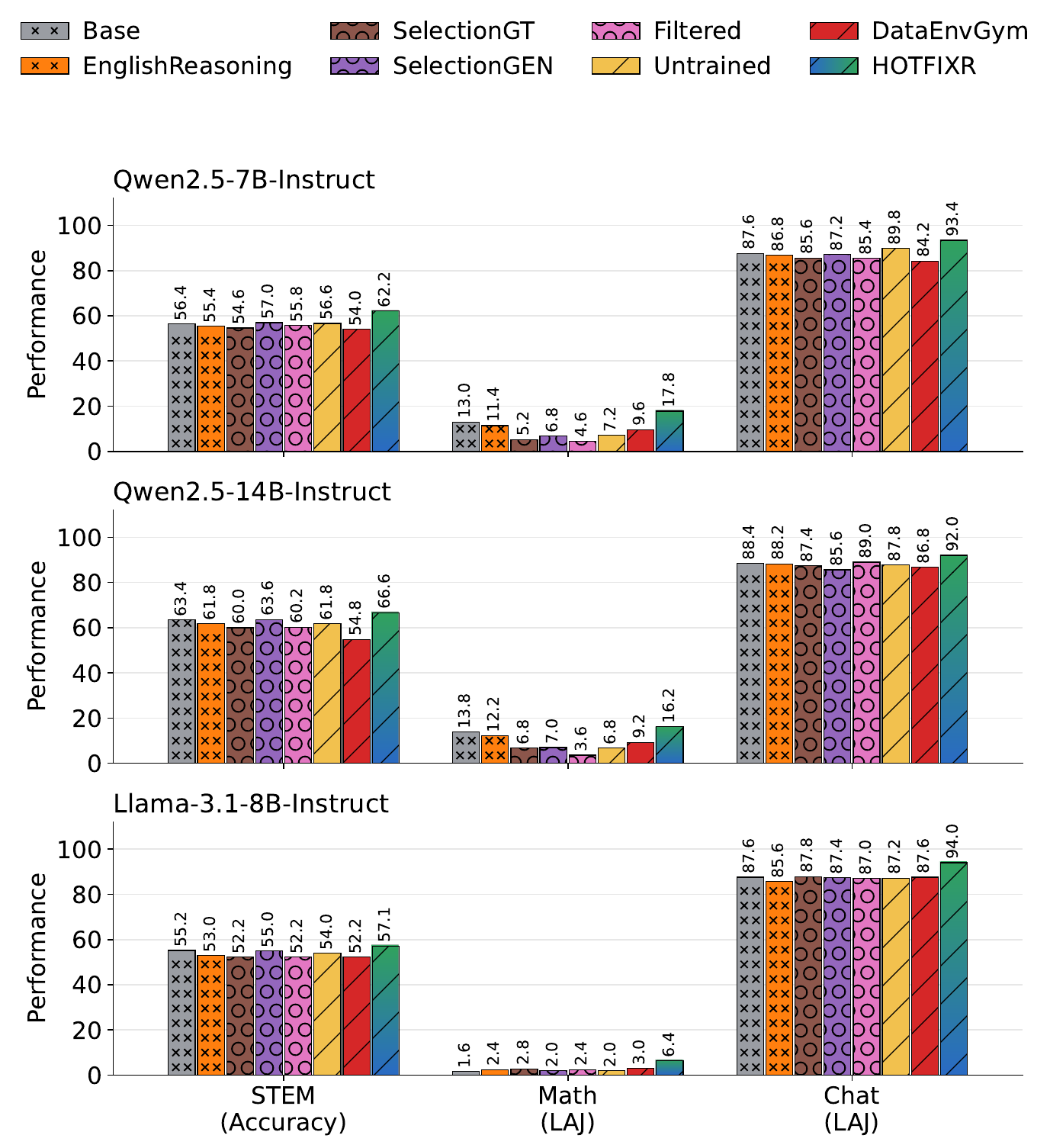}
    \caption{Performance of data curation methods \textbf{in-distribution}.}
    \label{fig: indistribution}
\end{figure}

Figure \ref{fig: indistribution} showcases the performance of all student models from baselines and \sysn for all three datasets, on the in-distribution task of Nemotron STEM, MATH, and CHAT. To clarify, in all the plots, the bars with $\times$ are the untrained baselines, the bars with circles are data selection baselines, and the bars with diagonal lines are data synthesis baselines.

\begin{figure}[h]
    \centering
    \includegraphics[width=\linewidth]{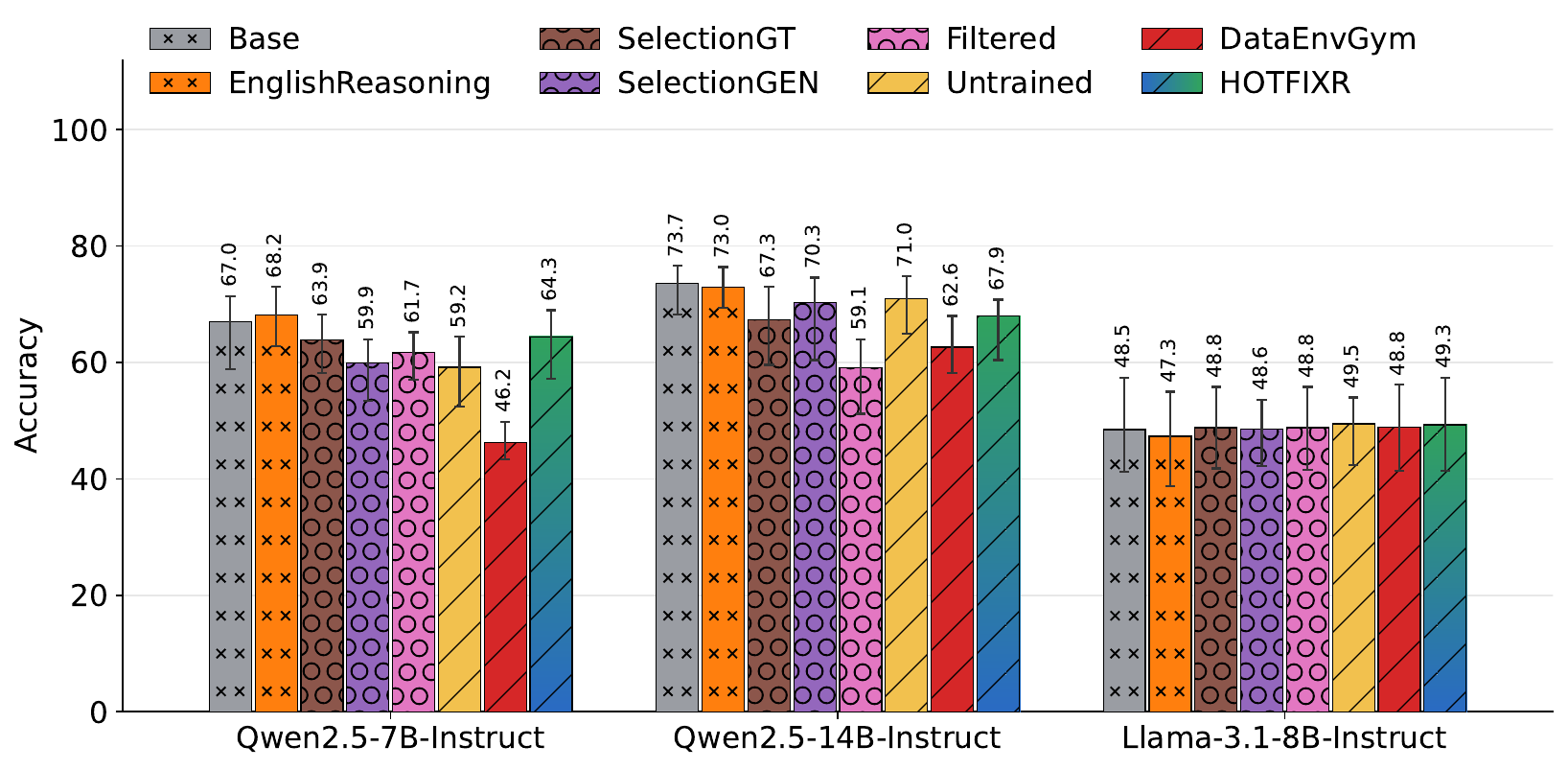}
    \caption{Performance of data curation methods for \textbf{factual (MMMLU)} queries, an out-of-distribution task.}
    \label{fig: factual}
\end{figure}

\begin{figure}[h]
    \centering
    \includegraphics[width=\linewidth]{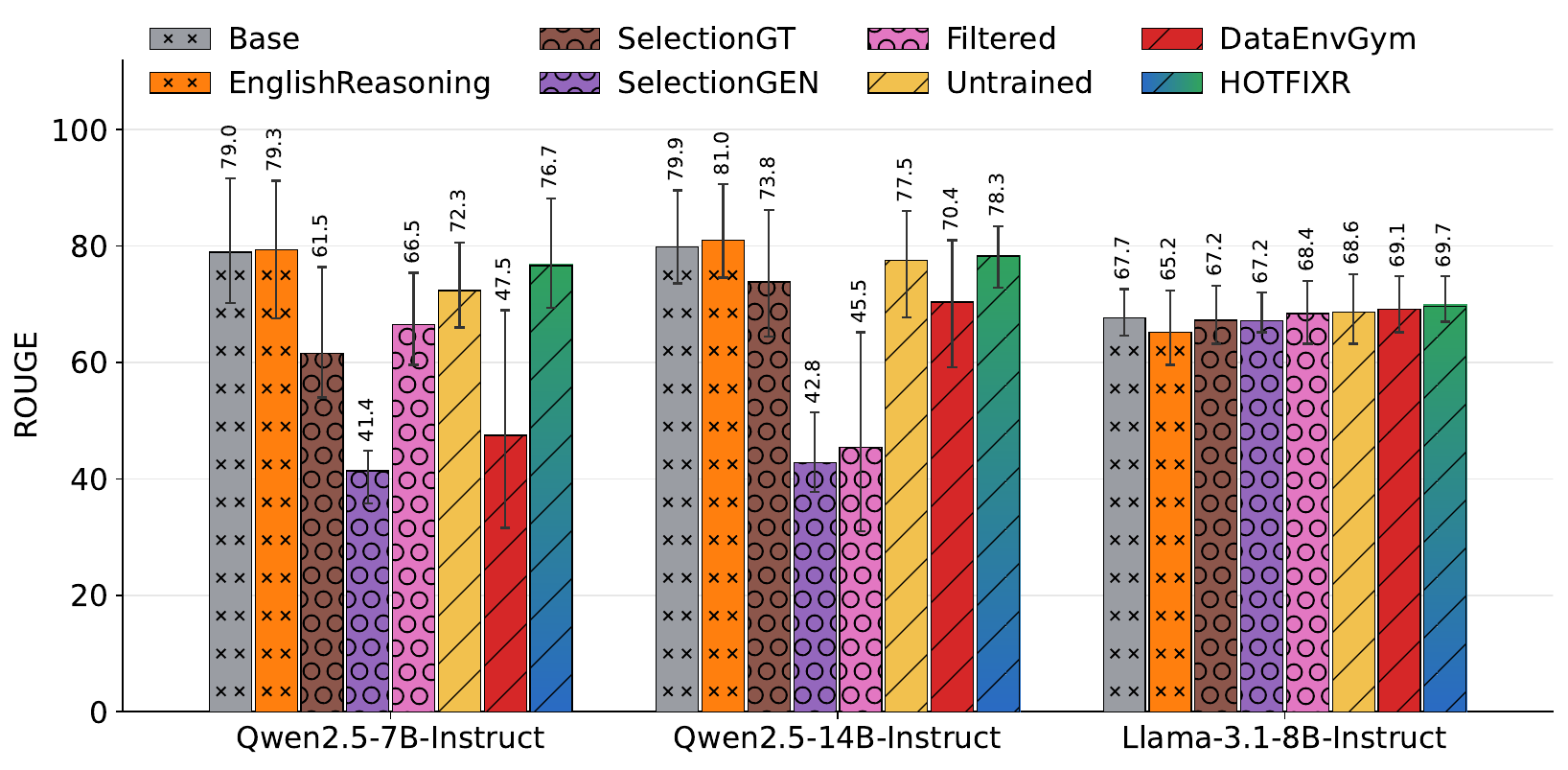}
    \caption{Performance of data curation methods for \textbf{RAG (multilingual HotPotQA)} queries, an out-of-distribution task.}
    \label{fig: rag}
\end{figure}

\begin{figure}[h]
    \centering
    \includegraphics[width=\linewidth]{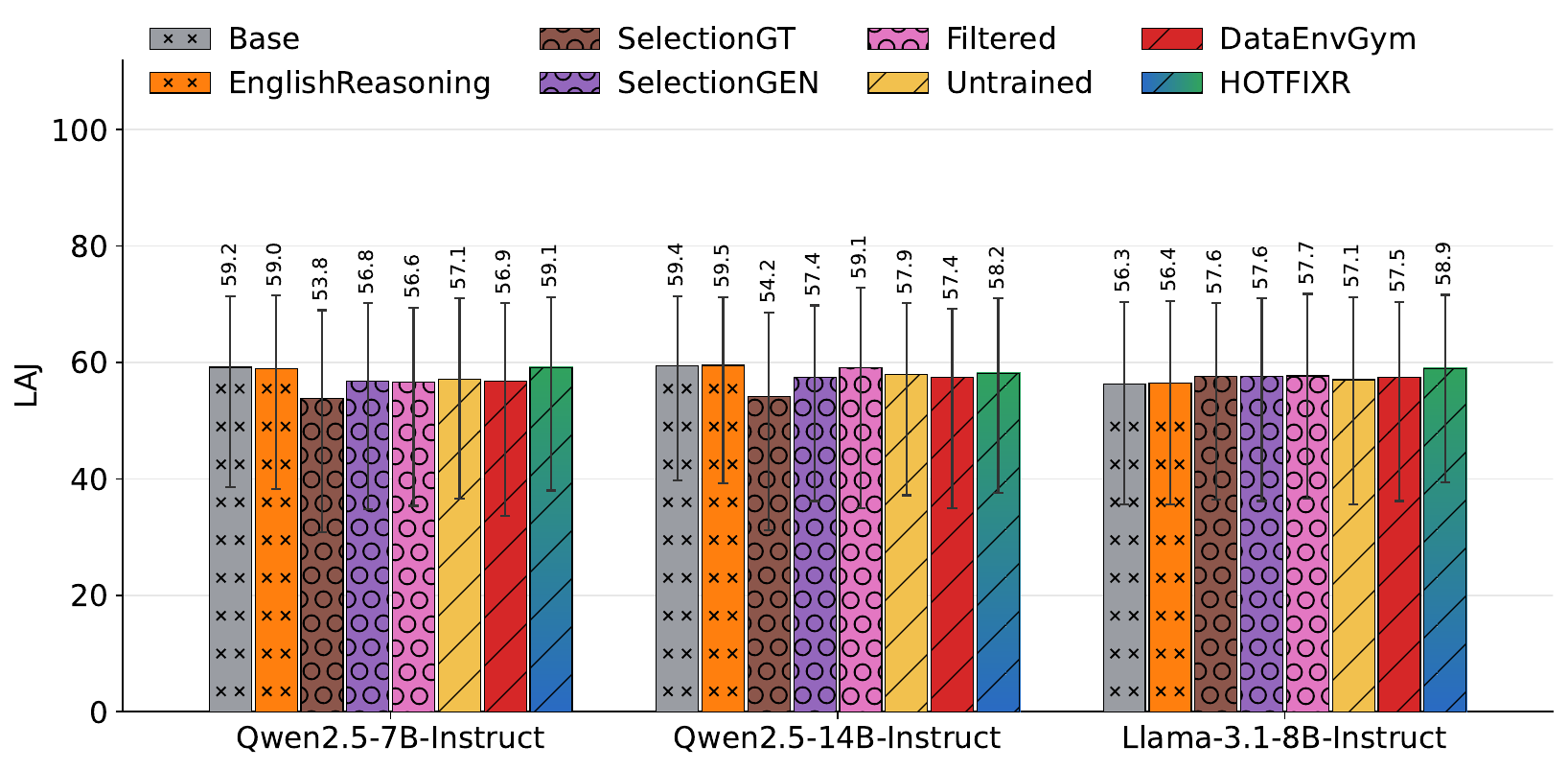}
    \caption{Performance of data curation methods for \textbf{translation (OPUS-100)} queries, an out-of-distribution task.}
    \label{fig: translation}
\end{figure}

Figures \ref{fig: factual}, \ref{fig: rag}, and \ref{fig: translation} compare the performance of all student models trained (or untrained) by all baselines on out-of-distribution tasks, including factual parametric knowledge, RAG, and translation, respectively. These plots have whiskers on each plot. For clarity, we report the average result across languages for each baseline. The whisker length indicates the minimum and maximum language performance. The shorter the whisker, the more consistent the performance is across languages. In Appendix \ref{app: per_language}, we report the granular per-language results for each task.

\subsection{Analysis}

\begin{takeaway}
 Takeaway 1: \sysn generates more informative in-distribution data.
\end{takeaway}
\vspace{3mm}
\noindent 
On Nemotron STEM, MATH, and CHAT tasks, \sysn gets an overall improvement of \textbf{~4.3\% over the base model} (5.5\% on Qwen 7B, 3.1\% on Qwen 14B, and 4.4\% on Llama 8B), improvement of \textbf{~6.0\% over the \textit{best} selection} method (7.5\% on Qwen 7B, 6.2\% on Qwen 14B, and 4.4\% on Llama 8B), and an improvement of \textbf{5.8\% over the \textit{best} synthesis} method (6.6\% on Qwen 7B, 6.1\% on Qwen 14B, and 4.8\% on Llama 8B).

\begin{table}[h]
\centering
\small
\begin{tabular}{lcccc}
\toprule
 & \multicolumn{2}{c}{In-Distribution} & \multicolumn{2}{c}{Out-of-Distribution} \\
\cmidrule(lr){2-3}\cmidrule(lr){4-5}
Baseline & $\Delta$ & Wins & $\Delta$ & Wins \\
\midrule
\Base & \plus4.3 & 9/9 & \minus0.9 & 3/9 \\
\EngReason& \plus5.4 & 9/9 & \minus0.7 & 4/9 \\
\SelGT & \plus7.0 & 9/9 & \plus3.8 & 9/9 \\
\SelGEN & \plus6.0 & 9/9 & \plus8.9 & 8/9 \\
\Filtered & \plus7.3 & 9/9 & \plus6.6 & 8/9 \\
\Untrained & \plus5.8 & 9/9 & \plus1.4 & 7/9 \\
\DataEnvGym & \plus7.1 & 9/9 & \plus7.3 & 9/9 \\
\midrule
\textbf{Average} & \textbf{\plus6.2} & \textbf{9/9} & \textbf{\plus3.7} & \textbf{7/9} \\
\bottomrule
\end{tabular}
\caption{Average performance difference of \sysn versus each baseline. The average is over nine settings: three students (Qwen 7B, Qwen 14B, Llama 8B) $\times$ three ID/OOD tasks. ``Wins'' counts the number of settings where \sysn{} scores higher.}
\label{tab: deltas}
\end{table}

\begin{takeaway}
Takeaway 2: \sysn minimizes out-of-distribution degradation.
\end{takeaway}
\vspace{3mm}
\noindent 
Because all baselines train on 5,000 samples, it is expected that the models will lose some OOD performance relative to the base model. There are two ways we evaluate for OOD generalization performance: by task, and by language.

 \textbf{By OOD task generalization}, we refer to the performance on Factual Knowledge (MMMLU), Translation (OPUS-100), and RAG (mHotPotQA). According to Table \ref{tab: deltas}, \sysn performance degrades by 0.9\% compared to the base model (averaged over three OOD tasks and three students). However, \textbf{\sysn is on average 5.6\% better than other training-based baselines}. Compared to \SelGT and \SelGEN, we achieve a 6.4\% improvement. This is because the questions are generic in the selection methods, and not targeted towards a model's weaknesses. The effect of \Untrained versus \sysn in OOD tasks is small but nontrivial (1.4\%), which shows that synthetic data will not harm OOD. Still, \sysn achieves a 5.8\% improvement in-distribution, so training data generation models is still important to achieve overall performance. Finally, over \DataEnvGym, \sysn performs 7.3\% better in OOD tasks. This is because \DataEnvGym is designed to generate data that mitigate student mistakes on a particular dataset. Instead, \sysn's dataset-agnostic reward is able to probe a student model's general weaknesses.

\begin{table}[h]
\centering
\small
\setlength{\tabcolsep}{5pt}
\begin{tabular}{lccccc}
\toprule
Method & De & Ja & Ru & Zh & Avg \\
\midrule
\SelGT      & -3.3 & -4.7 & -9.5 & -10.5 & -7.0 \\
\SelGEN     & \textbf{-1.9} & -4.0 & -22.5 & -21.5 & -12.5 \\
\Filtered   & -3.8 & -4.8 & -14.7 & -16.3 & -9.9 \\
\Untrained        & -2.9 & \un{-2.0} & \un{-2.3} & \un{-5.3} & \un{-3.1} \\
\DataEnvGym       & -6.8 & -6.0 & -12.2 & -15.4 & -10.1 \\ \midrule
\textbf{\sysn}   & \un{-2.6} & \textbf{-0.7} & \textbf{-0.9} & \textbf{-1.3} & \textbf{-1.4} \\
\bottomrule
\end{tabular}
\caption{OOD language generalization: performance differences relative to the \Base model, averaged across all models and corresponding tasks. Averaging the difference in \sysn and other methods, we can compute how much \sysn avoids catastrophic forgetting. For example, averaged across all languages, \sysn avoids catastrophic forgetting by 8.7\% compared to \DataEnvGym (10.1-1.4).}
\label{tab: ood-lang}
\end{table}

\textbf{By OOD language generalization}, we refer to the performance on languages not included in the training distribution. In Factual Knowledge (MMMLU) and Translation (OPUS-100), these are German and Japanese. In RAG (mHotPotQA), these are Russian and Chinese. Please refer to Appendix \ref{app: per_language} for the per-language results of all OOD tasks. Table \ref{tab: ood-lang} summarizes the results. Overall, there is post-training decline in performance. However, \sysn suffers the least: it performs better by 7.1\% compared to other training-based baselines.

\begin{table}[h]
\centering
\small
\setlength{\tabcolsep}{5pt}
\begin{tabular}{l cccccc}
\toprule
Method & Range & \begin{tabular}[c]{@{}c@{}}Trim.\\Range\end{tabular} & Std & IQR & CV \\
\midrule
\Base & 20.2 & 16.0 & 7.2 & 9.6 & \un{0.12} \\
\EngReason & 20.7 & 15.4 & 7.2 & 8.8 & 0.12 \\
\SelGT & 22.3 & 17.1 & 7.9 & 9.8 & 0.13 \\
\SelGEN & \textbf{18.8} & \un{14.9} & \textbf{6.9} & \un{9.5} & 0.13 \\
\Filtered & 22.6 & 17.1 & 8.0 & 10.6 & 0.14 \\
\Untrained & 20.1 & 15.0 & \un{7.1} & 9.7 & 0.12 \\
\DataEnvGym & 22.8 & 17.0 & 8.0 & 10.2 & 0.14 \\ \midrule
\sysn{} & \un{19.4} & \textbf{14.6} & \textbf{6.9} & \textbf{9.4} & \textbf{0.11} \\
\bottomrule
\end{tabular}
\caption{Cross-lingual spread of student performance, averaged over three student models (Qwen2.5-7B, Qwen2.5-14B, Llama-3.1-8B) and three multilingual tasks (OPUS, MMMLU, mHotpotQA) evaluated with each task's primary metric. Lower is more consistent for all dispersion columns; \textbf{Bold} marks the best value among data-generation methods. \textbf{Range} is the max $-$ min performance across all languages. \textbf{Trimmed Range} is the best minus the 2nd worst language accuracy. \textbf{Std} is standard deviation of the performance. \textbf{IQR} is the Inter-Quartile Range (difference in the 25\% and 75\% quartile). \textbf{CV} is the coeffiecient of variation (the standard devation divided by the mean).}
\label{tab: spread}
\end{table}

\begin{takeaway}
Takeaway 3: \sysn modestly improves multilingual consistency from the base model.
\end{takeaway}
\vspace{3mm}

\noindent 
Table \ref{tab: spread} reports various spread measures for determining the multilingual consistency of models. The \textbf{Range} is the max $-$ min performance across all languages. But, Range is not a reliant metric, as it suffers from outliers of low-resource languages. To avoid this outlier, we also report the \textbf{Trimmed Range}, which is the best minus the 2nd worst language accuracy. We also report the \textbf{Standard Deviation (Std)}, the \textbf{Inter-Quartile Range (IQR)}, and the \textbf{Coefficient of Variation (CV)}. All of these methods show that \sysn is slightly more consistent, but there is no SOTA data curation method for ensuring multilingual consistency. Still, we can say that \sysn, among other data curation methods, can maintain (and sometimes, improve) the multilingual consistency of LLMs.

\paragraph{Lingual Deficit score ablations.} Table \ref{tab: ablations} contains an ablation over the lingual deficit score components for training the question generator. We see that the format reward, LAI, and LSI do not have strong effects individually. But combined, they are able to train a strong data generator.

\paragraph{Knowledge Distillation.} In Appendix \ref{app: knowledge_distillation}, we test the effects of knowledge distillation (both questions and labels) from Qwen 32B. The results show that although it can improve performance, the majority of the gains come from the learned question generation in \sysn.

\paragraph{Continual Learning.} In Appendix \ref{app: continual_learning}, we train a fresh question generator on a continuously trained student, and see significant improvements of 7\% on Nemotron tasks, 8\% on OOD tasks, and 8\% on seen v.s. 7\% on unseen languages.


\begin{table}[t]
\centering
\small
\setlength{\tabcolsep}{4pt}
\begin{tabular}{llcccc}
\toprule
& & \multicolumn{1}{c}{ID} & \multicolumn{3}{c}{OOD} \\
\cmidrule(lr){3-3} \cmidrule{4-6}
Model & Config & Nemo. & Trans. & Fact. & RAG \\
\midrule
\multirow{4}{*}{Qwen 7B}
 & Format          & 52.0 & 57.6 & 64.0 & 75.8 \\
 & \;+ LAI         & 50.9 & 57.9 & 64.3 & 77.2 \\
 & \;+ LSI         & 50.8 & 57.8 & 63.6 & 77.7 \\
 & \;+ LAI + LSI   & 57.8 & 59.1 & 64.3 & 76.7 \\
\midrule
\multirow{4}{*}{Llama 8B}
 & Format          & 48.2 & 57.8 & 49.9 & 70.2 \\
 & \;+ LAI         & 48.5 & 57.2 & 49.2 & 69.2 \\
 & \;+ LSI         & 47.5 & 57.5 & 49.3 & 67.8 \\
 & \;+ LAI + LSI   & 52.5 & 58.9 & 49.3 & 69.7 \\
\midrule
\multirow{4}{*}{Qwen 14B}
 & Format          & 49.3 & 58.0 & 71.5 & 79.8 \\
 & \;+ LAI         & 50.9 & 58.6 & 66.3 & 76.0 \\
 & \;+ LSI         & 48.8 & 57.2 & 62.3 & 76.8 \\
 & \;+ LAI + LSI   & 58.3 & 58.2 & 67.9 & 78.3 \\
\bottomrule
\end{tabular}
\caption{Reward-signal ablation. All rows include the format reward. ``Nemo.'' is Nemotron, ``Trans.'' is Translation, ``Fact.'' is Factual. Highlights the impact each reward has---there is more effect of the reward functions combined, than individually. LAI is the ``Language Agnostic Incompetency'' and LSI is the ``Language Specific Incompetency'', as described in Section \ref{sec: method}.}
\label{tab: ablations}
\end{table}

\subsection{Discussion}
    In business use cases, where it is important to adapt a model to a particular task while also maintaining the model's general reasoning, context understanding, and answer generation abilities, \sysn offers the best of both worlds. Compared to the base model, it achieves a 4.3\% improvement, and loses only 0.9\% performance on out-of-distribution task. Ultimately, fine-tuned models will lose some generalization performance for a price. Across all the training-based baselines, \sysn achieves the best performance (6.2\% average improvement) and generalization (5.6\% average improvement).

Still, \sysn costs more due to the GRPO optimization of the data generator. With roughly 40 steps of RL optimization and 6 A100 NVIDIA GPUs, \sysn occurs a cost of 0.96min/training data sample (which is roughly 8 hours for $|\mathcal{D}_{QG}|=500$) to train a data generator. \textit{However, 8 GPU-hours is a one-time cost}. Once a model is trained for data generation, it can generate as much data as required. Furthermore, there are no efforts required to obtain high-quality seed data, clean generated data, or remove any PII content.

\begin{takeaway}
\sysn does increase training time, but it makes up for it with consistent improvements in distribution, and reliable generalization performance on out-of-distribution tasks.
\end{takeaway}
\vspace{3mm}

\section{Conclusion}
In this paper, we tackle the problem of adapting models to improve their multilingual abilities without degrading the overall performance. To do so, we present \sysn: a principled data synthesis pipeline that trains data generation models to generate data that targets the improvement of agentic abilities and the consistency in multilingual settings. Our reward for the data generator finds samples that are both difficult (LAI) and have larger multilingual representation gaps (LSI). In our experimentation, we see that \sysn is able to improve performance on in-distribution benchmarks, and remain reliable on OOD tasks and languages. Future work involves scaling experiments in the pretraining regime, where most of the multilingual representations are formed.

\section{Limitations}
Our work primarily focuses on high-resource languages: English, French, Spanish, German, Japanese, Portuguese, Italian. Even though we also test with Arabic, accounting for low-resource languages requires special consideration, that we aim to address in future work. For now, a demonstration of fine-tuning models to generate good data for high-resource languages is a nontrivial task in itself. Furthermore, with LLMs that face heavy training data biases, there is a risk of generating harmful data---we restrict our study to verifiable tasks that have clear correct and incorrect answers. We cannot predict whether our findings will extrapolate to unverifiable domains. Finally, our results rely on a strong label generation model that can output reliable labels (\texttt{Qwen/Qwen2.5-32B-Instruct}). In preliminary experiments, we tried generating labels with the respective student models themselves, but their instruction following abilities (in particular, ensuring that they generated labels in the prompted language) are poor. So we opted for generating labels with larger models. In spite of this weakness, we still showcase improved performance with few samples---in future work, we will explore a synthetic, active learning setup where models can generate questions that they want labeled, but also reduce the amount of data used. This can help ensure the labeling cost (either by LLM or by human) is reduced.

\bibliography{custom}

\appendix
\onecolumn
\section{Reward Hacking}
\label{app: reward_hacking}
In Section 3.2, we mention that our training dataset $D_{QG}$ is only 500 because the question generator reward hacks. We would like to clarify what that means.

Although Figure \ref{fig: acquisition} is an illustrative example, it can still help us understand reward hacking. When a data generation model reward hacks, it finds the sample with the local maximum lingual deficit reward. It will not explore further and gets stuck there. During inference, this results in the data generator generating that one sample (with the local maximum reward) over and over again. Even if we change the in-context sample, the generator will generate the exact same data point repeatedly. This means that the resulting dataset will be 5,000 copies of the exact same sample, which will cause trained student models to fail catastrophically. Hence, to avoid reward hacking, we simply reduce the number of optimization steps in GRPO (to around 41 steps) and reduce the number of training samples.

\clearpage
\section{Error Bars for Performance on ID and OOD tasks, \textit{and} Per-Language Performance on OOD tasks.}
\label{app: per_language}
\begin{figure*}[h]
    \centering
    \includegraphics[width=0.8\linewidth]{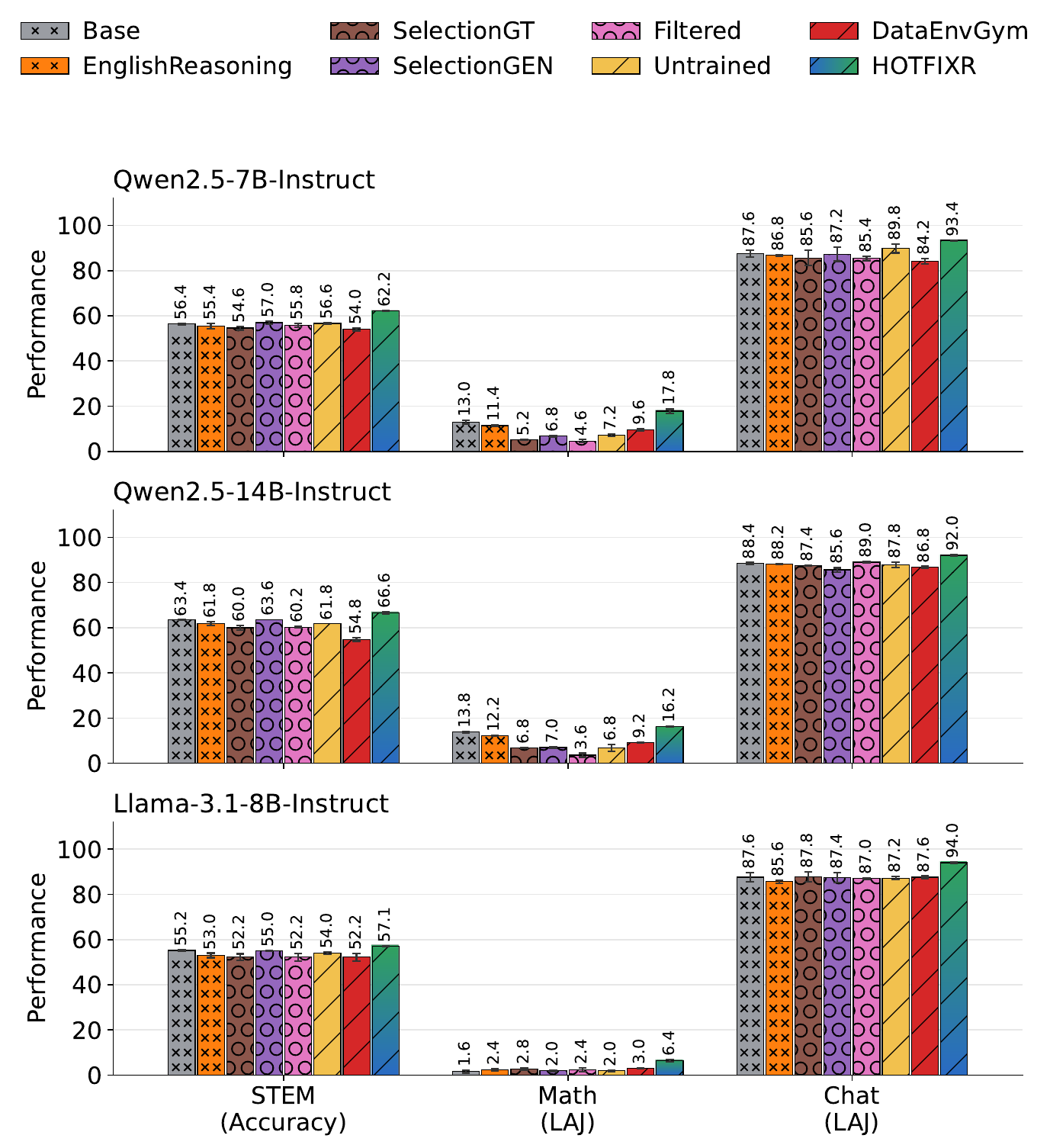}
    \caption{Performance of data curation methods for \textbf{Nemotron} queries, per task.}
    \label{fig: nemotron_appendix}
\end{figure*}

\begin{figure*}[h]
    \centering
    \includegraphics[width=0.8\linewidth]{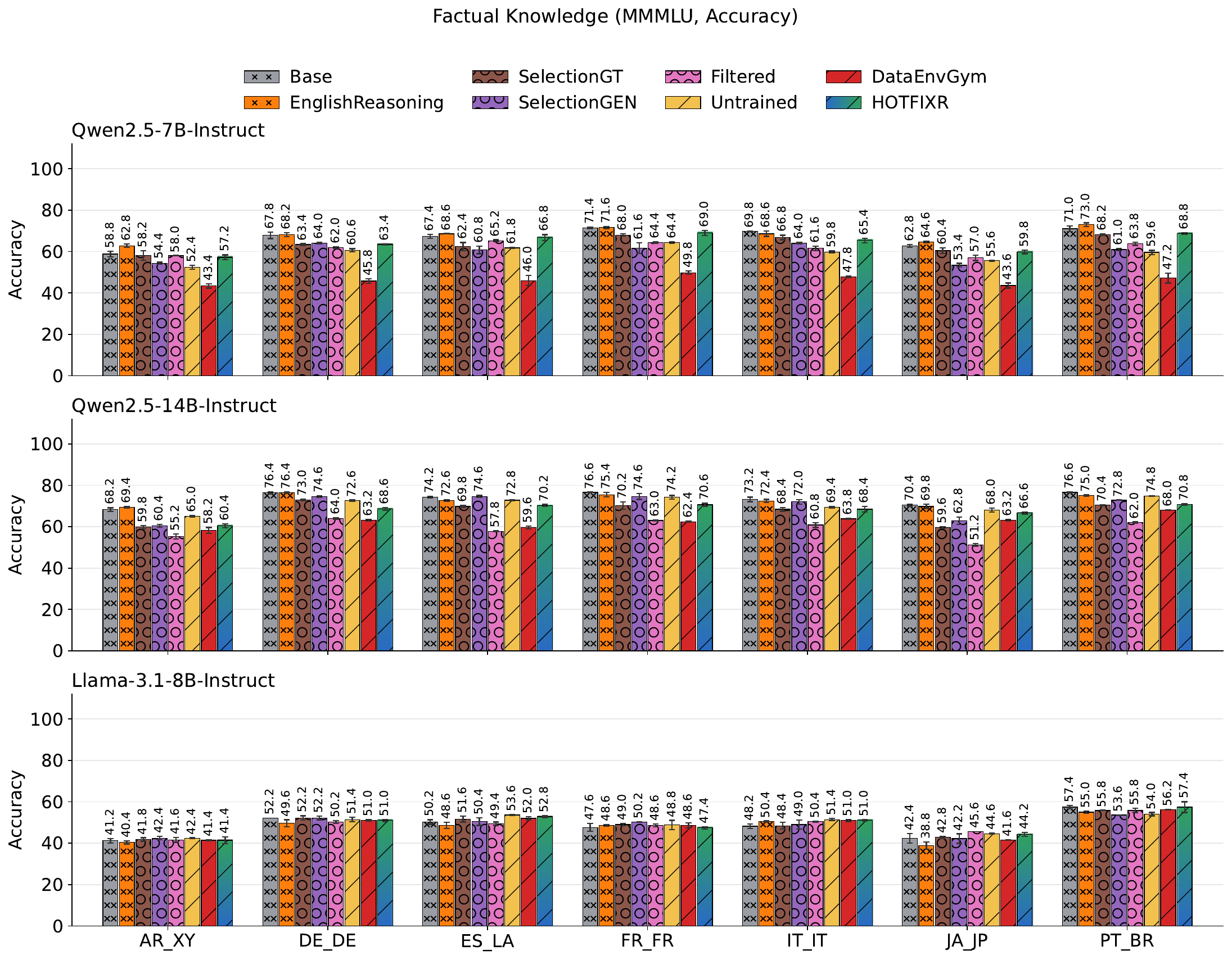}
    \caption{Performance of data curation methods for \textbf{factual (MMMLU)} queries, per language.}
    \label{fig: perlang_factual}
\end{figure*}

\begin{figure*}[h]
    \centering
    \includegraphics[width=0.8\linewidth]{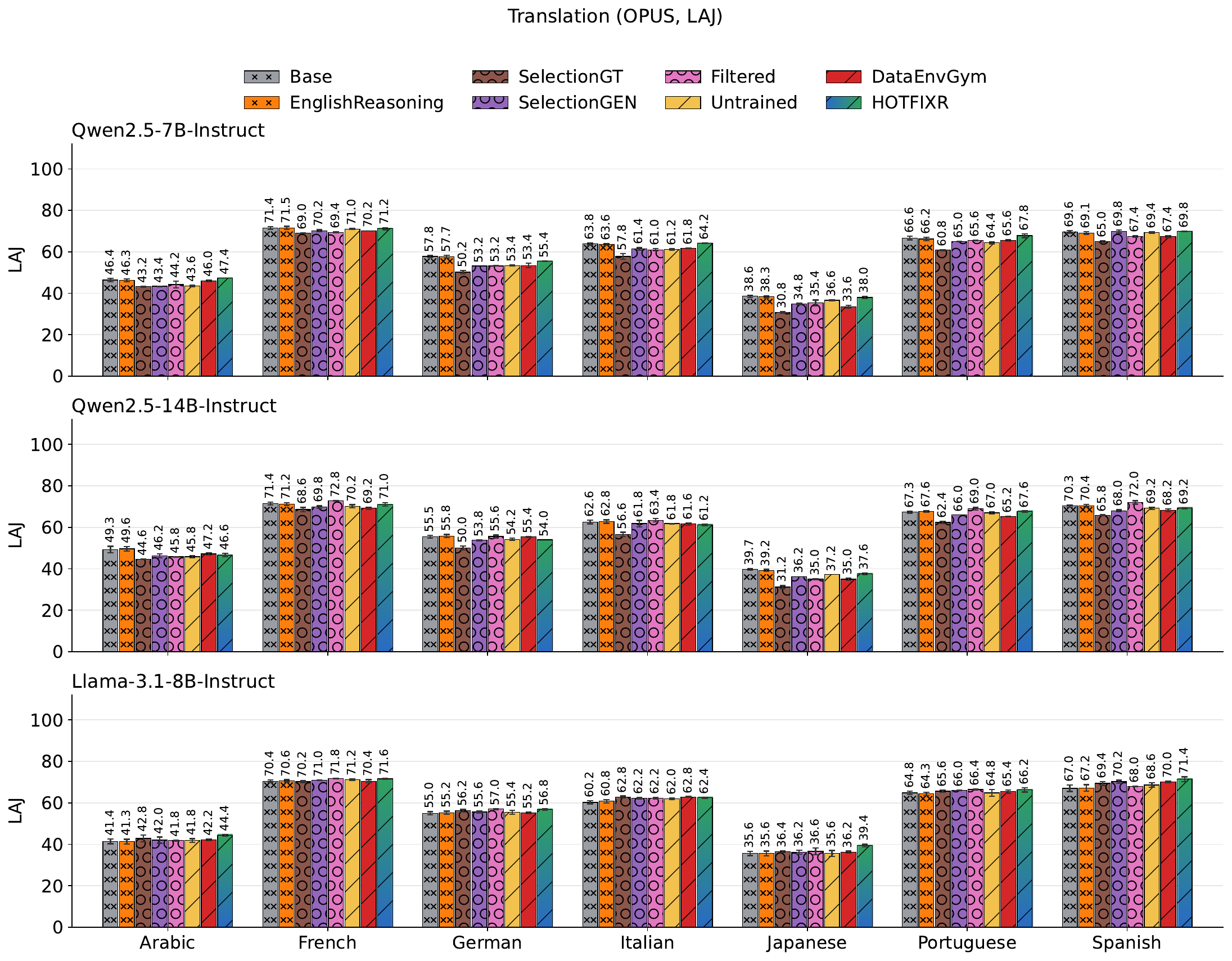}
    \caption{Performance of data curation methods for \textbf{translation (OPUS-100)} queries, per language.}
    \label{fig: perlang_translation}
\end{figure*}

\begin{figure*}[h]
    \centering
    \includegraphics[width=\linewidth]{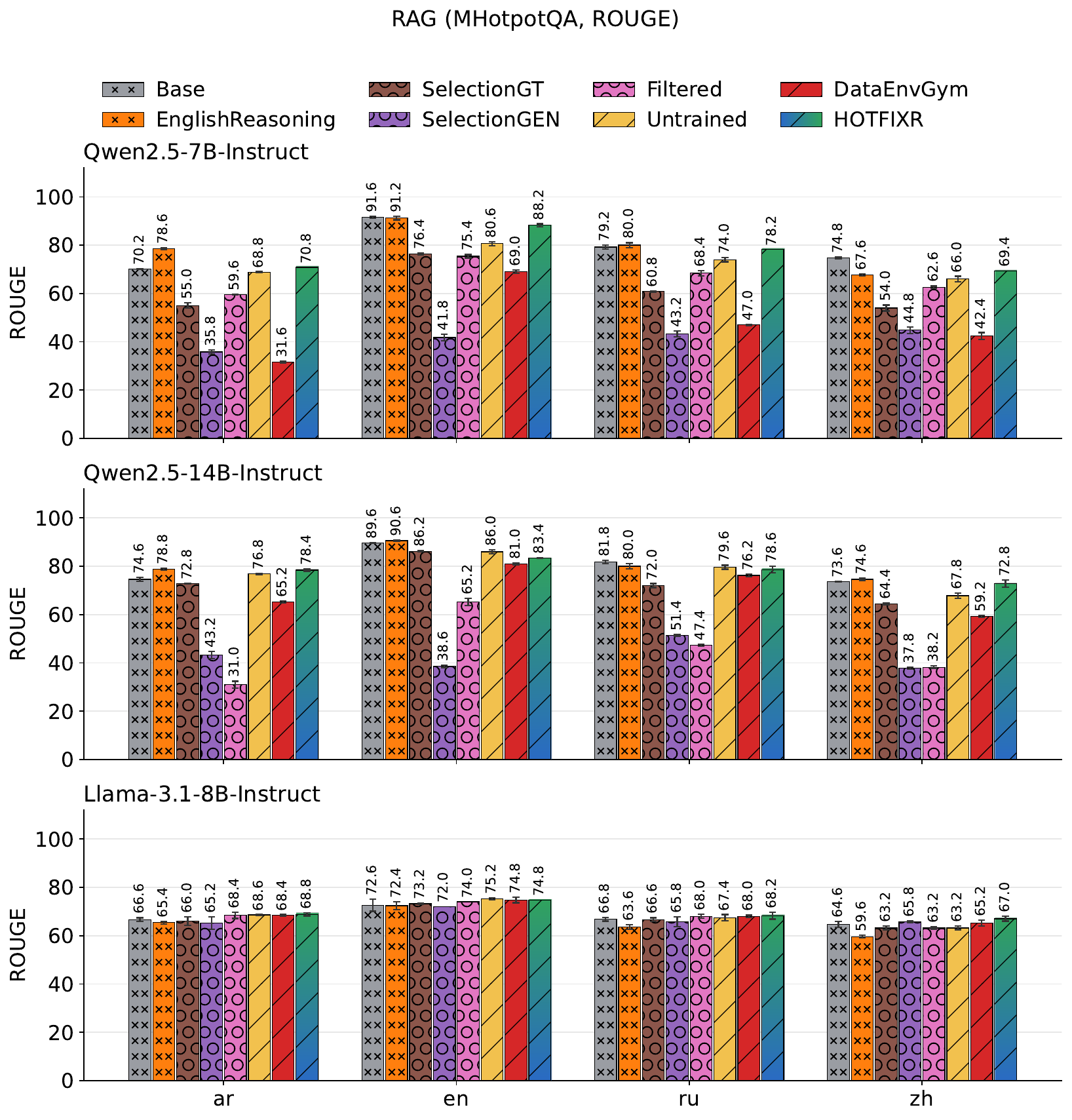}
    \caption{Performance of data curation methods for \textbf{RAG (multilingual HotPotQA)} queries, per language.}
    \label{fig: perlang_rag}
\end{figure*}

Figure \ref{fig: nemotron_appendix} contains the error bars for all Nemotron tasks. Figures \ref{fig: perlang_factual}, \ref{fig: perlang_translation} and \ref{fig: perlang_rag} contain the error bars as well as the per-language performance of all methods, for each model on the factual, translation, and RAG tasks, respectively.

Because the error bars are quite small on the figures themselves, we also note down the minimum and maximum standard deviations for each subtask for the Nemotron (Table \ref{tab: std_nemo}), Translation (Table \ref{tab: std_trans}), Factual (Table \ref{tab: std_fact}) and RAG (Table \ref{tab: std_rag})) tasks.

\begin{table}[h]\centering\small
\begin{tabular}{lcccccc}
\toprule
 & \multicolumn{2}{c}{Qwen 7B} & \multicolumn{2}{c}{Qwen 14B} & \multicolumn{2}{c}{Llama 8B} \\
\cmidrule(lr){2-3}\cmidrule(lr){4-5}\cmidrule(lr){6-7}
Method & min & max & min & max & min & max \\
\midrule
\Base & 0.30 & 1.50 & 0.10 & 0.40 & 0.20 & 2.00 \\
\EngReason & 0.25 & 1.23 & 0.16 & 0.75 & 0.34 & 1.00 \\
\SelGT & 0.20 & 3.40 & 0.10 & 1.00 & 0.40 & 2.20 \\
\SelGEN & 0.30 & 3.10 & 0.00 & 0.90 & 0.10 & 2.20 \\
\Filtered & 0.70 & 0.90 & 0.30 & 0.80 & 0.30 & 1.70 \\
\Untrained & 0.40 & 1.90 & 0.00 & 1.50 & 0.30 & 0.70 \\
\DataEnvGym & 0.40 & 1.20 & 0.10 & 0.70 & 0.20 & 1.70 \\
\sysn & 0.10 & 0.90 & 0.20 & 0.40 & 0.05 & 0.50 \\
\bottomrule
\end{tabular}
\caption{Minimum and maximum standard deviation (across 3 runs) for the Nemotron task (STEM, MATH, and CHAT), per model and method. This explains the small error bars in Figure \ref{fig: nemotron_appendix}.}
\label{tab: std_nemo}
\end{table}

\begin{table}[h]\centering\small
\begin{tabular}{lcccccc}
\toprule
 & \multicolumn{2}{c}{Qwen 7B} & \multicolumn{2}{c}{Qwen 14B} & \multicolumn{2}{c}{Llama 8B} \\
\cmidrule(lr){2-3}\cmidrule(lr){4-5}\cmidrule(lr){6-7}
Method & min & max & min & max & min & max \\
\midrule
\Base & 0.43 & 0.85 & 0.41 & 1.55 & 0.58 & 1.56 \\
\EngReason & 0.40 & 0.83 & 0.43 & 1.05 & 0.55 & 1.56 \\
\SelGT & 0.10 & 1.40 & 0.20 & 1.00 & 0.40 & 1.60 \\
\SelGEN & 0.00 & 0.80 & 0.00 & 1.50 & 0.10 & 1.50 \\
\Filtered & 0.10 & 1.60 & 0.10 & 0.80 & 0.10 & 1.60 \\
\Untrained & 0.20 & 0.50 & 0.00 & 0.70 & 0.30 & 1.70 \\
\DataEnvGym & 0.00 & 1.10 & 0.20 & 0.70 & 0.20 & 0.80 \\
\sysn & 0.00 & 0.90 & 0.00 & 0.80 & 0.20 & 1.10 \\
\bottomrule
\end{tabular}
\caption{Minimum and maximum standard deviation (across 3 runs) for the Translation task (OPUS-100), per model and method. This explains the small error bars in Figure \ref{fig: perlang_translation}.}
\label{tab: std_trans}
\end{table}

\begin{table}[h]\centering\small
\begin{tabular}{lcccccc}
\toprule
 & \multicolumn{2}{c}{Qwen 7B} & \multicolumn{2}{c}{Qwen 14B} & \multicolumn{2}{c}{Llama 8B} \\
\cmidrule(lr){2-3}\cmidrule(lr){4-5}\cmidrule(lr){6-7}
Method & min & max & min & max & min & max \\
\midrule
\Base & 0.10 & 1.50 & 0.10 & 1.10 & 0.00 & 2.30 \\
\EngReason & 0.09 & 1.27 & 0.28 & 1.09 & 0.33 & 1.73 \\
\SelGT & 0.30 & 2.30 & 0.20 & 1.80 & 0.10 & 1.60 \\
\SelGEN & 0.40 & 2.60 & 0.10 & 1.80 & 0.10 & 2.30 \\
\Filtered & 0.10 & 1.30 & 0.20 & 1.20 & 0.00 & 1.20 \\
\Untrained & 0.30 & 1.00 & 0.10 & 1.00 & 0.20 & 2.20 \\
\DataEnvGym & 0.30 & 2.40 & 0.10 & 1.40 & 0.10 & 1.20 \\
\sysn & 0.30 & 1.40 & 0.30 & 1.30 & 0.30 & 2.60 \\
\bottomrule
\end{tabular}
\caption{Minimum and maximum standard deviation (across 3 runs) for the Factual task (MMMLU), per model and method. This explains the small error bars in Figure \ref{fig: perlang_factual}.}
\label{tab: std_fact}
\end{table}

\begin{table}[h]\centering\small
\begin{tabular}{lcccccc}
\toprule
 & \multicolumn{2}{c}{Qwen 7B} & \multicolumn{2}{c}{Qwen 14B} & \multicolumn{2}{c}{Llama 8B} \\
\cmidrule(lr){2-3}\cmidrule(lr){4-5}\cmidrule(lr){6-7}
Method & min & max & min & max & min & max \\
\midrule
\Base & 0.10 & 0.90 & 0.10 & 0.80 & 0.80 & 2.50 \\
\EngReason & 0.33 & 0.99 & 0.16 & 1.05 & 0.53 & 1.71 \\
\SelGT & 0.30 & 1.20 & 0.10 & 0.80 & 0.30 & 1.80 \\
\SelGEN & 0.80 & 1.40 & 0.30 & 1.50 & 0.00 & 2.50 \\
\Filtered & 0.10 & 1.10 & 0.40 & 1.50 & 0.00 & 1.30 \\
\Untrained & 0.20 & 1.10 & 0.30 & 1.10 & 0.10 & 1.30 \\
\DataEnvGym & 0.20 & 1.40 & 0.30 & 0.60 & 0.30 & 1.20 \\
\sysn & 0.00 & 0.60 & 0.20 & 1.50 & 0.00 & 1.40 \\
\bottomrule
\end{tabular}
\caption{Minimum and maximum standard deviation (across 3 runs) for the RAG task (mHotPotQA), per model and method. This explains the small error bars in Figure \ref{fig: perlang_rag}.}
\label{tab: std_rag}
\end{table}

\clearpage
\section{Knowledge Distillation}
\label{app: knowledge_distillation}
In order to separate the gains from the question generation versus Qwen 32B's label generation, we use Qwen 32B to generate both the questions and labels. This is similar to the \Untrained baseline, but with Qwen 32B. Table \ref{tab: kd} contains those results, in which we call the new baseline ``Distillation (32b)''. The table shows that while distillation with a larger model does help improve performance, \sysn's trained question generation model is able to understand model weaknesses much better than simply distilling knowledge from a larger model.

\begin{table}[h]\centering\small
\begin{tabular}{lcccc}
\toprule
Method & Nemotron & Factual & Translation & RAG \\
\midrule
\multicolumn{5}{c}{\texttt{Qwen2.5-7B-Instruct}} \\ \midrule
\Base & 52.3 & 67.0 & 59.2 & 79.0 \\
\EngReason & 51.2 & 68.2 & 59.0 & 79.3 \\
\SelGT & 48.5 & 63.9 & 53.8 & 61.5 \\
\SelGEN & 50.3 & 59.9 & 56.8 & 41.4 \\
\Filtered & 48.6 & 61.7 & 56.6 & 66.5 \\
mCOT & 47.9 & 55.3 & 54.4 & 53.8 \\
\Untrained & 51.2 & 59.2 & 57.1 & 72.3 \\
\DataEnvGym & 49.3 & 46.2 & 56.9 & 47.5 \\
Distillation (32B) & 49.6 & 64.7 & 58.9 & 77.3 \\
\sysn & 57.8 & 64.3 & 59.1 & 76.7 \\
\midrule
\multicolumn{5}{c}{\texttt{Qwen2.5-14B-Instruct}} \\ \midrule
\Base & 55.2 & 73.7 & 59.4 & 79.9 \\
\EngReason & 54.1 & 73.0 & 59.5 & 81.0 \\
\SelGT & 51.4 & 67.3 & 54.2 & 73.8 \\
\SelGEN & 52.1 & 70.3 & 57.4 & 42.8 \\
\Filtered & 50.9 & 59.1 & 59.1 & 45.5 \\
\Untrained & 52.1 & 71.0 & 57.9 & 77.5 \\
\DataEnvGym & 50.3 & 62.6 & 57.4 & 70.4 \\
Distillation (32B) & 53.3 & 69.5 & 59.0 & 78.4 \\
\sysn & 58.3 & 67.9 & 58.2 & 78.3 \\
\midrule
\multicolumn{5}{c}{\texttt{Llama-3.1-8B-Instruct}} \\ \midrule
\Base & 48.1 & 48.5 & 56.3 & 67.7 \\
\EngReason & 47.0 & 47.3 & 56.4 & 65.2 \\
\SelGT & 47.6 & 48.8 & 57.6 & 67.2 \\
\SelGEN & 48.1 & 48.6 & 57.6 & 67.2 \\
\Filtered & 47.2 & 48.8 & 57.7 & 68.4 \\
\Untrained & 47.7 & 49.5 & 57.1 & 68.6 \\
\DataEnvGym & 47.6 & 48.8 & 57.5 & 69.1 \\
Distillation (32B) & 47.7 & 51.1 & 58.0 & 70.3 \\
\sysn & 52.5 & 49.3 & 58.9 & 69.7 \\
\bottomrule
\end{tabular}
\caption{Average performance per model, with the added Distillation method. Here, we are able to show that a lot of \sysn's empirical success comes from the targeted question generation, rather than having correct, distilled labels from a large model.}
\label{tab: kd}
\end{table}

\clearpage
\section{An Experiment on Continual Learning}
\label{app: continual_learning}

\begin{figure}[h]
    \centering
    \includegraphics[width=0.49\linewidth]{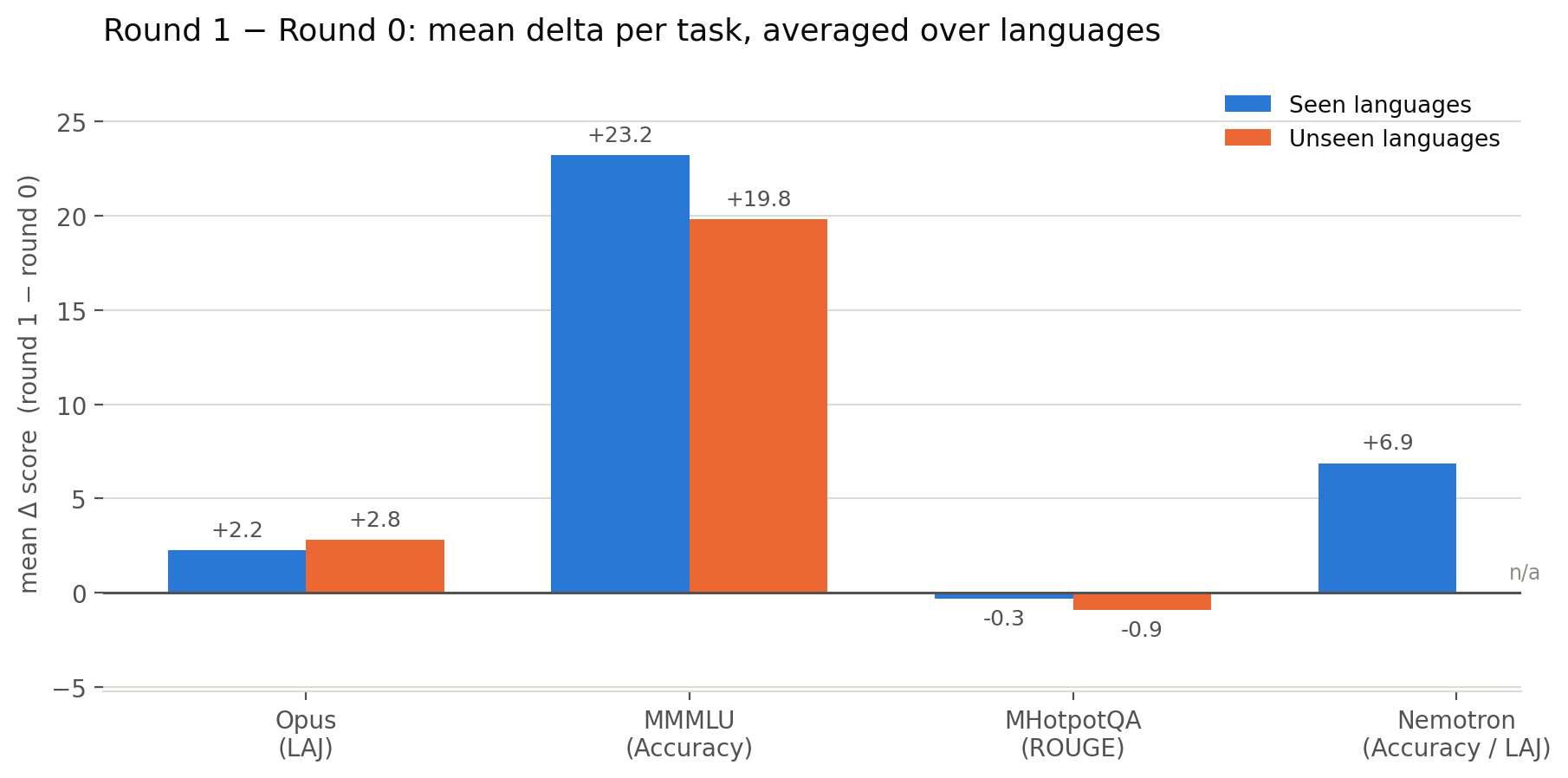}
    \includegraphics[width=0.49\linewidth]{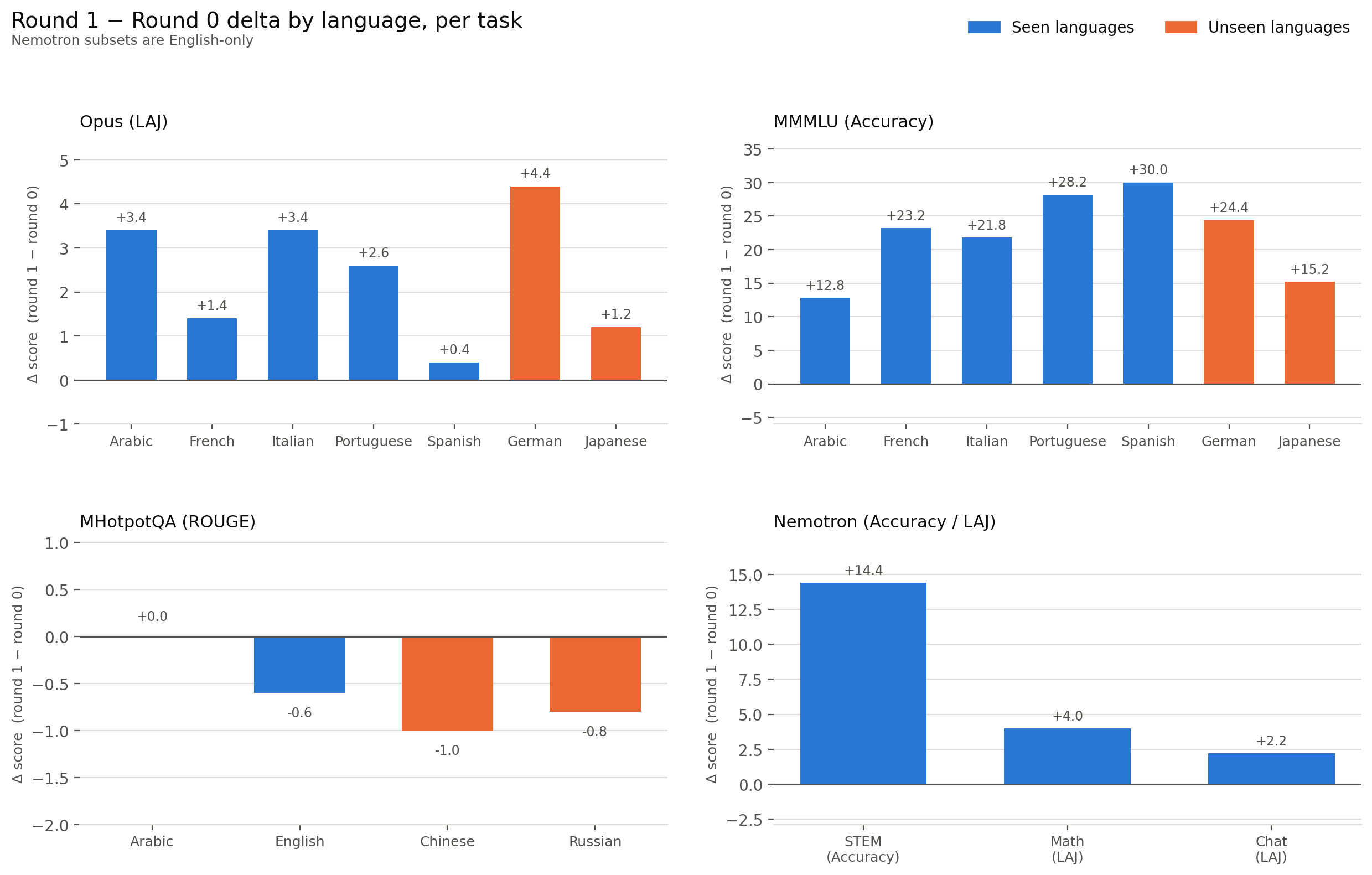}
    \caption{The delta in performance of student models trained with two rounds of \sysn versus one round. The improvement with two rounds is much more than the improvement with one round.}
    \label{fig: rounds}
\end{figure}

In this section, we describe an experiment where we test the effects of a second round of \sysn. The question generator model is trained from scratch (the base model), but the student model is continuously trained (from the checkpoint of Round 0 of \sysn). We call the first iteration Round 0 and the second iteration (of a fresh generator model, but continuously trained student model) Round 1. In Figure \ref{fig: rounds}, we present the aggregate results and the per-task results of the experiment, respectively on the left and right. This experiment is only on the Qwen 7B setting.

As shown, even though the question generation model is trained from scratch, each iteration of \sysn improves the student model much more significantly, even on unseen, OOD tasks and languages. These improvements show that using \sysn in iterations will yield higher gains not only in cross-linguality but also model response quality.

\clearpage
\section{Prompts}
\label{app: prompts}
Figures \ref{prompt: math}-\ref{prompt: chat} contain the three prompts used for \sysn's question generator training and data generation for each subset of Nemotron (MATH, STEM, and CHAT).

\begin{figure*}[h]
  \centering
  \begin{tcolorbox}[
    colback=gray!5!white,
    colframe=black!75!black,
    title=Prompt for \sysn training on Nemotron MATH,
    boxrule=0.3mm,
    width=\textwidth,
    arc=1.5mm,
    auto outer arc
  ]

\begin{verbatim}
In {language}, generate a NEW, ORIGINAL math problem that is AS DIFFICULT as the 
reference below. Do NOT copy, paraphrase, or reuse it in any way.

Reference (difficulty calibration only — do not reproduce):
<question>
{question}
</question>

<reasoning>
{reasoning}
</reasoning>

<answer>
{answer}
</answer>

Requirements for your generated problem:
- The question, reasoning, and answer should all be in {language}.
- Requires non-trivial reasoning steps (no single-step shortcuts)
- Draws from: number theory, combinatorics, algebra, geometry, or probability
- Is self-contained and precisely stated
- Reasoning should include a complete step-by-step derivation
- Answer includes just the final result

IMPORTANT: generate a (question, reasoning, answer) triplet; wrap your question,
reasoning, and answer in the following special tokens:
<question> Insert your {language} question here. </question>
<reasoning> Insert the thinking and general reasoning here in {language}. 
   </reasoning>
<answer> Insert your short {language} answer here. </answer>
\end{verbatim}

  \end{tcolorbox}
  \caption{Prompt used during \sysn training and data generation, for the Nemotron MATH subset. As input, the prompt takes a \texttt{question}, \texttt{answer}, and \texttt{reasoning} from the Nemotron MATH training dataset, and a \texttt{language} that alternates between English, French, Spanish, Arabic, Portuguese, or Italian.}
  \label{prompt: math}
\end{figure*}

\begin{figure*}[h]
  \centering
  \begin{tcolorbox}[
    colback=gray!5!white,
    colframe=black!75!black,
    title=Prompt for \sysn training on Nemotron STEM,
    boxrule=0.3mm,
    width=\textwidth,
    arc=1.5mm,
    auto outer arc
  ]

\begin{verbatim}
In {language}, generate a NEW, ORIGINAL STEM multiple-choice question (MCQA) that 
is AS DIFFICULT as the reference below. Do NOT copy, paraphrase, or reuse it in 
any way.

Reference (difficulty calibration only — do not reproduce):
<question>
{question}
</question>

<reasoning>
{reasoning}
</reasoning>

<answer>
{answer}
</answer>

Requirements for your generated question:
- The question, reasoning, and answer should all be in {language}.
- Draws from STEM domains: physics, chemistry, biology, computer science, 
  engineering, or mathematics
- Requires non-trivial conceptual or quantitative reasoning (no single-step 
  lookups or trivial recall)
- Has exactly 4 answer choices labeled (A), (B), (C), (D) — only one is correct
- Distractors are plausible and reflect common misconceptions or near-miss 
  reasoning errors
- Is self-contained, unambiguous, and precisely stated
- Reasoning walks through the correct derivation/justification step by step and 
  explains why each distractor is wrong
- Answer is the correct letter only, e.g. "(B)"

IMPORTANT: generate a (question, reasoning, answer) triplet; wrap them in the 
following special tokens:
<question> Insert your {language} question stem followed by the four answer 
  choices (A)–(D). </question>
<reasoning> Insert the step-by-step reasoning in {language}, including why each 
  distractor is incorrect. </reasoning>
<answer> Insert only the correct letter, e.g. "(A)". </answer>
\end{verbatim}

  \end{tcolorbox}
  \caption{Prompt used during \sysn training and data generation, for the Nemotron STEM subset. As input, the prompt takes a \texttt{question}, \texttt{answer}, and \texttt{reasoning} from the Nemotron STEM training dataset, and a \texttt{language} that alternates between English, French, Spanish, Arabic, Portuguese, or Italian.}
  \label{prompt: stem}
\end{figure*}

\begin{figure*}[h]
  \centering
  \begin{tcolorbox}[
    colback=gray!5!white,
    colframe=black!75!black,
    title=Prompt for \sysn training on Nemotron CHAT,
    boxrule=0.3mm,
    width=\textwidth,
    arc=1.5mm,
    auto outer arc
  ]

\begin{verbatim}
In {language}, generate a NEW, ORIGINAL instruction-following task that is AS 
COMPLEX as the reference below. Do NOT copy, paraphrase, or reuse it in any way.

Reference (complexity calibration only — do not reproduce):
<question>
"{question}"
</question>

<reasoning>
{reasoning}
</reasoning>

<answer>
{answer}
</answer>

Requirements for your generated task:
- The question, reasoning, and answer should all be in {language}.
- Instruction imposes at least as many explicit constraints as the reference (e.g. 
    format, length, style, content restrictions, conditional logic)
- Constraints are specific and verifiable — a reader can check whether the 
    response satisfies each one
- Draws from: open-ended knowledge tasks (Alpaca-style), conversational requests 
    (LMArena-style), or format-constrained tasks (IFEval-style)
- Instruction is self-contained and unambiguous
- Reasoning walks through how each constraint is satisfied, step by step
- Answer is a complete response that fully obeys every constraint in the instruction

IMPORTANT: generate a (question, answer) pair; wrap your question and answer in 
the following special tokens:
<question> Insert your {language} instruction here. </question>
<reasoning> Insert the step-by-step reasoning in {language}. </reasoning>
<answer> Insert the complete response in {language} that satisfies all 
    constraints. </answer>
\end{verbatim}

  \end{tcolorbox}
  \caption{Prompt used during \sysn training and data generation, for the Nemotron CHAT subset. As input, the prompt takes a \texttt{question}, \texttt{answer}, and \texttt{reasoning} from the Nemotron CHAT training dataset, and a \texttt{language} that alternates between English, French, Spanish, Arabic, Portuguese, or Italian.}
  \label{prompt: chat}
\end{figure*}

\section{LLM Usage}
In writing this paper, the only main use of LLMs was for creating figures, specifically Claude Opus 4.8. After providing the experimental results from our results, the authors prompted it to write code using matplotlib to create the figures. Other LLM usages were very minimal: only \textit{enhancing} the writing and small code suggestions. We did not use LLMs to write code files, perform literature reviews, or describe our methodology/evaluation. All LLM use was reviewed heavily by the authors.

\newpage

\end{document}

%% file: gradient_font.tex
\usepackage{xcolor}
\usepackage{xspace}

\definecolor{sysBlue}{HTML}{2273B4}
\definecolor{sysGreen}{HTML}{47B571}

\ExplSyntaxOn
\NewDocumentCommand{\gradienttext}{ m m m }
 {
  \sys_gradient_text:nnn { #1 } { #2 } { #3 }
 }

\int_new:N \l_sys_len_int
\int_new:N \l_sys_pos_int
\fp_new:N \l_sys_mix_fp

\cs_new_protected:Npn \sys_gradient_text:nnn #1 #2 #3
 {
  \int_set:Nn \l_sys_len_int { \tl_count:n { #3 } }
  \int_zero:N \l_sys_pos_int
  \tl_map_inline:nn { #3 }
   {
    \int_incr:N \l_sys_pos_int
    \int_compare:nNnTF { \l_sys_len_int } > { 1 }
     {
      \fp_set:Nn \l_sys_mix_fp
       { 100 * ( \l_sys_len_int - \l_sys_pos_int ) / ( \l_sys_len_int - 1 ) }
     }
     { \fp_set:Nn \l_sys_mix_fp { 100 } }
    \textcolor{ #1 ! \fp_to_decimal:N \l_sys_mix_fp ! #2 }{ ##1 }
   }
 }
\ExplSyntaxOff